\documentclass[fleqn,10pt,twocolumn]{SWARM25}

\usepackage{amsmath}
\usepackage{amssymb}
\usepackage{color}
\usepackage[normalem]{ulem}
\usepackage{comment}

\title{Ant-inspired Walling Strategies for Scalable Swarm Separation:
Reinforcement Learning Approaches Based on Finite State Machines}

\author{Shenbagaraj Kannapiran${}^{1}$, Elena Oikonomou${}^{2}$, Albert Chu${}^{2\dagger}$, Spring Berman${}^{1}$, and \\ Theodore P.~Pavlic${}^{2,3}$}
\speaker{Albert Chu}

\affils{%
${}^{1}$School for Engineering of Matter, Transport and Energy, Arizona State University, Tempe, AZ 85287, USA\\
(E-mail: \texttt{shenbagaraj@asu.edu}, \texttt{spring.berman@asu.edu})\\
${}^{2}$School of Computing and Augmented Intelligence, Arizona State University, Tempe, AZ 85281, USA\\
(E-mail: \texttt{e.oikonomou@asu.edu}, \texttt{abchu@asu.edu}, \texttt{tpavlic@asu.edu})\\
${}^{3}$School of Life Sciences, Arizona State University, Tempe, AZ 85281, USA\\
}

\abstract{%
In natural systems, emergent structures often arise to balance competing demands. Army ants, for example, form temporary ``walls'' that prevent interference between foraging trails. Inspired by this behavior, we developed two decentralized controllers for heterogeneous robotic swarms to maintain spatial separation while executing concurrent tasks. The first is a finite-state machine~(FSM)-based controller that uses encounter-triggered transitions to create rigid, stable walls. The second integrates FSM states with a Deep Q-Network~(DQN), dynamically optimizing separation through emergent ``demilitarized zones.'' In simulation, both controllers reduce mixing between subgroups, with the DQN-enhanced controller improving adaptability and reducing mixing by 40--50\% while achieving faster convergence.%
}


\begin{document}

\pagestyle{headings}

\maketitle

%%%%%%%%%%%%%%%%%%%%%%%%%%%%%%%%%%%%%%%%%%%%%%%%%%%%%%%%%%%%%%%%%%%%%%%%%%%%%%%%
\section{INTRODUCTION}

Autonomous spatial coordination %in large-scale robotic swarms 
remains a fundamental challenge in swarm robotics, particularly in scenarios where heterogeneous subgroups must operate concurrently without mutual interference. %In many applications, %Whether in disaster zones, agricultural fields, or warehouse environments, 
Swarm effectiveness often depends on the robots' ability %of robots 
to maintain task-specific spatial allocations while dynamically adapting to environmental and inter-agent interactions. A critical functionality in such contexts is the decentralized separation of subgroups, ensuring that they %different collectives 
can perform their respective tasks without obstructing each other. For example, swarms engaged in disaster response %missions 
could %employ walling behaviors to 
form protective barriers to create corridors for safe transport of victims to medical assistance teams. In warehouses, robots with depleted energy can benefit from unobstructed pathways to recharging stations created by walling off fully charged robots, significantly reducing interference and thus decreasing their average time %required 
to recharge. In swarms performing agricultural tasks, %walling behaviors 
subgroup separation can facilitate dedicated harvesting routes, prevent collisions, and maintain efficient task distribution among specialized groups of robots.

\begin{figure}
    \centering
    \includegraphics[width=\linewidth]{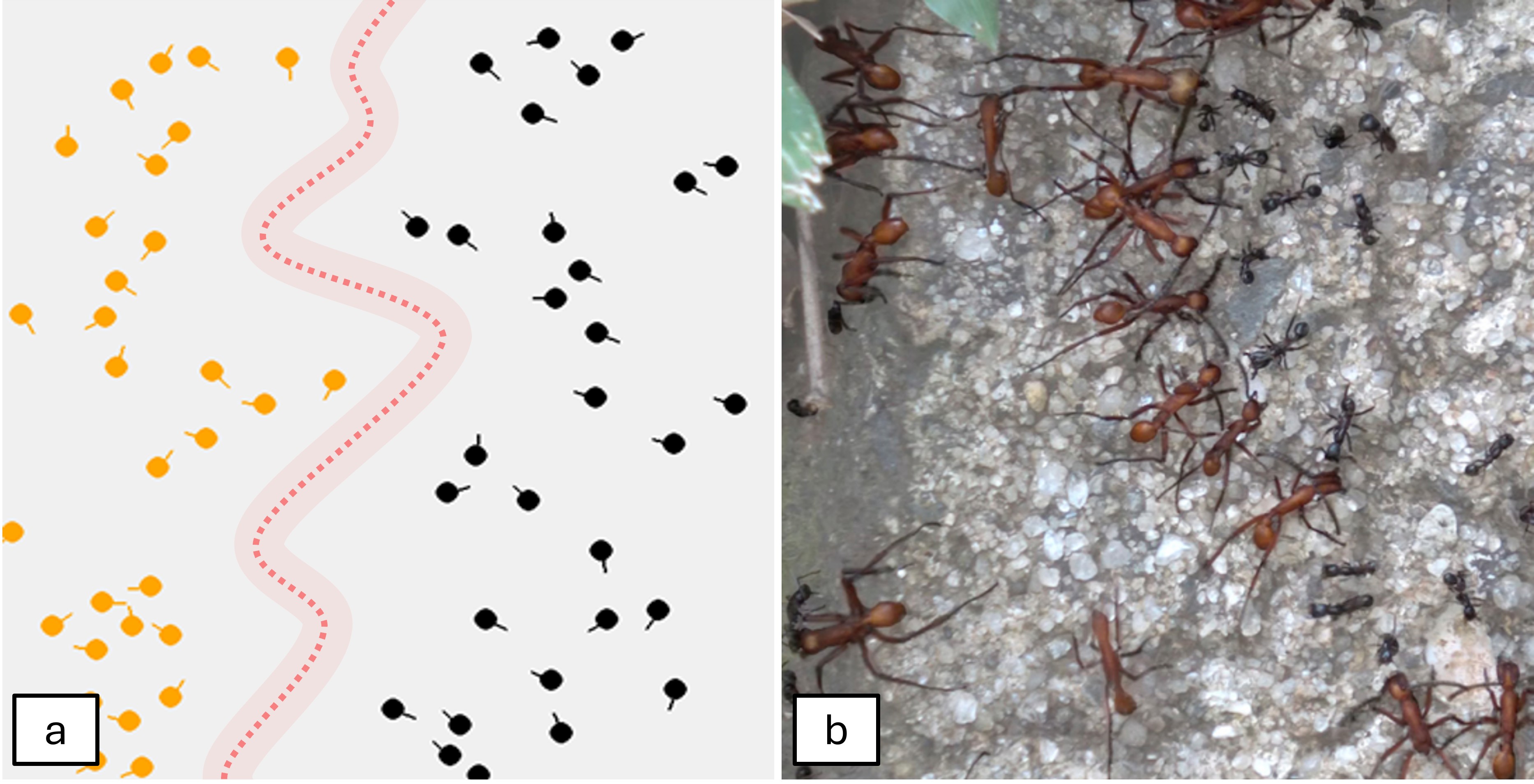}
    \caption{%Algorithm for Infrastructures that are Dissipative (AID) concept (not a working example). 
    Illustration of example ``dissipative infrastructures'' that enforce self-organized separation in collectives. (a) Simulation snapshot %Example desired output in a simulated arena 
    showing populations of two types of agents (\emph{orange, black}) that maintain separation by following one of the controllers presented in this paper, resulting in the emergence of a dynamic ``virtual wall'' (\emph{dotted line}) between the populations.
    %by remaining on either side of a virtual wall
    %forming a virtual wall to maintain separation between two groups. 
    (b) Inspiration from observations in Baudier and Pavlic~\cite{baudier2020incidental} of self-organized walls of army ants that act to prevent two large trails from intersecting.}
    \label{fig:WallingPics}
\end{figure}

To address this challenge, we focus on the design of robot controllers that produce ``dissipative infrastructures'': temporary, emergent structures formed by subgroups of robots that reduce interference by acting as physical or virtual barriers~(Fig.~\ref{fig:WallingPics}(a)). These infrastructures facilitate the unimpeded, efficient operation of %performance of a primary task by
%safe and efficient operation of 
the larger swarm 
%unimpeded performance of a primary task
by enforcing local safety invariants, such as maintaining separation between subgroups or clearing pathways for time-sensitive operations. The controllers implement decentralized decision-making based on local sensing, temporarily reallocating robots to a support task (here, acting as a barrier) that maintains a safety invariant, thereby improving the swarm’s performance at the primary task.  
Our approach is inspired by natural systems, notably ant colonies, where certain ants dynamically shift from primary foraging roles to qualitatively different supportive behaviors like bridge-building~\cite{reid2015army} or wall-formation~(Fig.~\ref{fig:WallingPics}(b)). In particular, we draw on observations of ant walling behavior, where individuals temporarily halt and form stationary barriers %formations 
to redirect conflicting foraging trails, effectively minimizing interference between colonies. This bio-inspired mechanism 
%\spr{This type of ``self-sacrificing'' bio-inspired mechanism %is \spr{referred to as {\it self-organized sacrifice} in \cite{ramshanker2024strategic}, in which it 
%has been utilized in \cite{ramshanker2024strategic} to reduce the computational burden of cooperative localization in a robotic swarm inspection task. Here, we employ such a mechanism to enable} 
offers a compelling template for enabling decentralized, task-responsive spatial segregation in robotic swarms.

%we introduce a novel cooperative localization mechanism that minimizes collective computation expenditure through selforganized sacrifice. Here, a few agents bear the computational burden of localization; through local interactions, they improve the inspection productivity of the swarm

% In contrast, our controllers are designed to produce emergent ``dissipative infrastructures,'' comprised of physical or virtual barriers, that minimize interference between two subgroups of a swarm to facilitate their unimpeded performance of a primary task. The controllers implement decentralized decision-making based on local sensing, temporarily reallocating robots to a support task (here, acting as a barrier) that maintains a safety invariant, thereby improving the swarm’s performance at the primary task. 

%For example, within large, conspicuous foraging trails of army ants, small groups of ants can change from foraging to bridge-building behavior to use their own bodies to construct temporary shortcuts for nestmates to travel upon~\cite{reid2015army}. Furthermore, when two massive foraging trails from different colonies come into close proximity, some ants on both sides will organize into stationary wall formations~(Fig.~\ref{fig:WallingPics}(b)) that cause the trails behind them to move away from each other, at which point the walls naturally dissipate and those ants return to the foraging trails~\cite{baudier2020incidental}.

Various techniques for self-organized separation of subgroups within robotic swarms have been based on the differential adhesion hypothesis in cellular developmental biology, which posits that cells with different adhesion strengths tend to separate from each other. 
For example, decentralized artificial potential-based controllers based on this concept~\cite{kumar2010segregation} lead robots to experience different potentials when they interact with robots of different types. \emph{Differential potentials}~\cite{santos2014segregation,santos2020spatial} have also been developed to sort a heterogeneous swarm into spatially separated clusters and concentric rings of different robot types in both two and three dimensions. Swarm separation has also been achieved by mimicking cellular adhesion with magnetic attractive forces between robots~\cite{pan2024physical}.

Other controllers for spatial separation of heterogeneous swarms are %designed for robots with minimal capabilities that react to local information. 
minimal sensing-based strategies that rely on simple, local rules and sensor inputs to produce emergent group-level patterns.
In \cite{mitrano2019minimalistic}, a swarm of differential-drive robots self-organizes into an arbitrary number of groups using line-of-sight sensors that detect the presence and type of a nearby robot. 
% decentralized algorithm to achieve segregation into an arbitrary number of groups with swarms of autonomous robots. The distinguishing feature of our approach is in the minimalistic assumptions on which it is based. Specifically, we assume that (i) Each robot is equipped with a ternary sensor capable of detecting the presence of a single nearby robot, and, if that robot is present, whether or not it belongs to the same group as the sensing robot;
% studying the minimal assumptions for N -class segregation to emerge from local, decentralized interactions among robots. The term ‘N class’ refers to the creation of N spatially distinct groups. We show that, for segregation to emerge, it is sufficient to equip an ‘if/then/else’, differential drive robot with a ternary sensor. This sensor detects the presence of a robot along a line-of-sight. When a robot is detected, the sensor can distinguish whether it is a kin, i.e., it belongs to the same group as the sensing robot, or a non-kin, i.e., it belongs to a different group. When multiple robots are in range, the sensor returns information on the closest one of them.
The work~\cite{krischanski2023multi} also presents a differential-drive robot controller for multi-group aggregation that maps readings from a line-of-sight sensor to wheel velocities, and a deterministic global search algorithm was used to determine the controller parameters. 
The work~\cite{feola2022adaptive} achieves swarm robotic allocation between two types of collaborative tasks, each requiring a certain number of robots working in the same area, with a Finite-State Machine~(FSM)-based controller in which random-walk and dwell-time parameters are adapted in response to local task requirements and task collaboration outcomes. 

This prior work largely focuses on designing controllers that produce the specific behavior of swarm spatial separation as an end in itself, rather than as an adaptive support structure for ongoing primary tasks. However, recent work~\cite{ramshanker2024strategic} explores task-aware self-sacrifice behaviors to improve inspection productivity, illustrating how decentralized robot swarms can reallocate agents to support roles. Inspired by this emerging perspective, our work introduces a framework of ``dissipative infrastructures'' that generalizes beyond self-sacrificing behaviors.

This paper is organized as follows. In Section~\ref{sec:controllers}, we discuss two different approaches for designing swarm-based robot controllers for wall-type dissipative infrastructures. First, in Section~\ref{sec:FSM}, we introduce a hand-designed FSM-based robot controller to produce swarm separation behaviors similar to ant walling behaviors. The controller includes encounter-based state transitions, which use local sensor information about other robots, between random-exploration and stationary states. %, both with and without time delays in the stationary states.
Then, in Section~\ref{sec:RL}, we introduce
a hybrid RL--FSM framework that combines the behaviors modeled by the FSM controller with deep reinforcement learning~(RL). This integration leverages discrete states for low-level navigation and a novel DQN architecture for high-level behavior switching, creating a robust architecture that adapts to dynamic environments while maintaining swarm separation. Our DQN architecture processes three critical input modalities---distance measurements from (e.g.)~Ultra-Wideband~(UWB) sensor data~\cite{AlarifiEtAl16} for robust swarm navigation, Angle of Arrival~(AoA) data, and discrete indicators of swarm membership---enabling decentralized decision-making informed by rich spatial--contextual awareness. To efficiently handle the variable-sized input inherent in scenarios with large numbers of agents, our DQN implementation incorporates an attention mechanism that preserves linear computational complexity. This sensor-driven approach makes our solution particularly suitable for real-world deployments where conventional positioning systems may be unreliable. 
We describe our empirical approach to evaluating these two controllers in Section~\ref{sec:simulation}, and in Section~\ref{sec:results}, 
we demonstrate in simulation %empirically 
that our RL--FSM framework %successfully 
maintains stable spatial coverage, low mixing ratios, and linear computational scaling across a range of swarm sizes with sensor noise and communication constraints. We give concluding remarks and future directions in Section~\ref{sec:conclusion}.

\section{Robot Controllers}
\label{sec:controllers}

In this work, we focus on the design of swarm-based robot controllers have the potential for spontaneous emergence of dissipative infrastructures. In particular, we focus on two different approaches for the emergence of walling that enables \textbf{two co-existing swarms} to \textbf{maximally cover an area} while also \textbf{preventing intermixing}. Analogous to ant colonies, we designate a ``Nestmate'' of a robot as another robot of its type, and a ``Non-Nestmate'' as a robot of the other type. 

\subsection{Finite State Machine (SM) Controller} %Architecture} 
\label{sec:FSM}
%Drawing inspiration from army ant colonies, we implement a decentralized decision-making system where robots can dynamically switch between primary and secondary behaviors. The core mechanism is implemented as a finite state machine where agents transition between Moving and Walling states based on local sensing information.

% \[
% \begin{aligned}
% &\delta(\text{Moving}, \text{NonNestmateEncounter}) = \text{Walling} \\
% &\delta(\text{Moving}, \text{NestmateEncounter}) = \text{Moving}\\
% &\delta(\text{Walling}, \text{WallingTimerExpired}) = \text{AvoidNonNestmate} \\
% &\delta(\text{Walling}, \text{NestmateEncounter}) = \text{Walling} \\
% &\delta(\text{AvoidNonNestmate}, \text{AboveSafeDist}) = \text{Moving} \\
% &\delta(\text{AvoidNonNestmate}, \text{BelowSafeDist}) = \text{AvoidNonNestmate}.
% \end{aligned}
% \]

% \begin{itemize}
%     \item $Q = \{\text{Moving}, \text{Walling}, \text{AvoidNonNestmate}\}$ is the set of states.
%     \item $\Sigma = \{\text{NestmateEncounter}, \text{NonNestmateEncounter}, \text{WallingTimerExpired}, \text{AboveSafeDist}, \text{BelowSafeDist}, \text{NestmateEncounterMoving}\}$ is the input alphabet.
%     \item $q_0 = \text{Moving}$ is the initial state.
%     \item $F = \emptyset$ is the set of accepting states (not applicable here).
%     \item $\delta$ is the transition function:
% \end{itemize}

\begin{figure}
    \centering
    \includegraphics[width=0.9\linewidth]{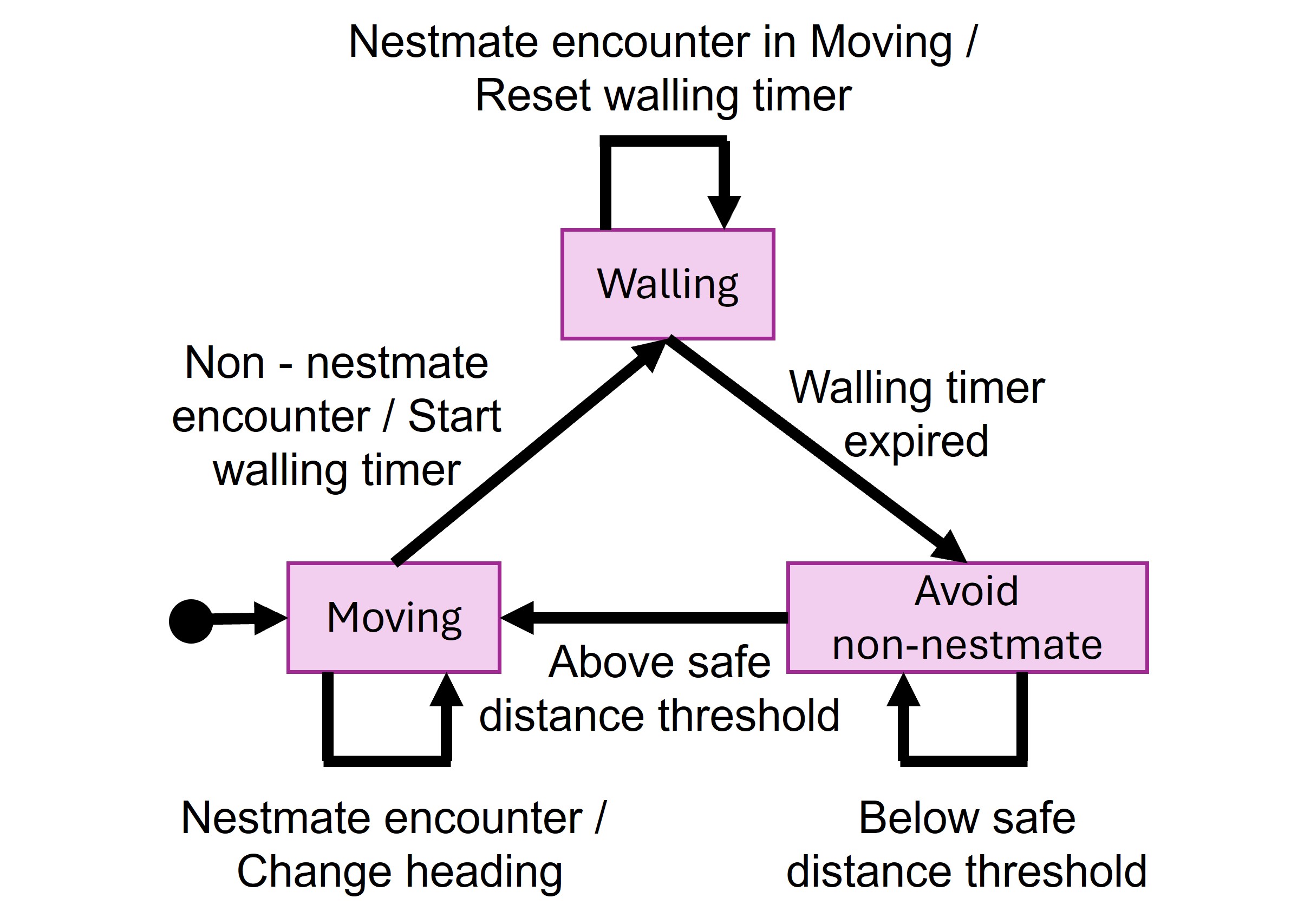}
    \vspace{2mm}
    \caption{Finite state machine (FSM)-based robot controller.}
    \label{fig:fsm}
\end{figure}
First, we introduce a Finite-State Machine~(FSM)-based controller mimicking the observed behaviors of real army ants engaged in walling~\cite{baudier2020incidental}. As illustrated in Fig.~\ref{fig:fsm}, it is defined by the 5-tuple $(Q, \Sigma, \delta, q_0, F)$ where:
\begin{itemize}
    \item $Q = \{\textit{Walling}, \textit{Moving}, \textit{Avoid Non-Nestmate}\}$ is the set of states;
    \item $\Sigma = \{\text{NestmateEncounter},~ \text{NonNestmateEncounter},~ \\
    \text{WallingTimerExpired},~ \text{AboveSafeDist},~ \text{BelowSafeDist},~ \\
    \text{MovingNestmateEncounter}
    \}$ is the set of input symbols;
    \item $q_0 = \textit{Moving}$ is the initial state;
    \item $F = \emptyset$ is the set of accepting states (not applicable here); and
    \item $\delta$ is the transition function capturing the edges of Fig.~\ref{fig:fsm}.
\end{itemize} \vspace{1mm}

%In the Finite State Machine (FSM)-based controller, 
A robot begins in the \textit{Moving} state, in which it follows a correlated random walk~(CRW), e.g., to perform an exploration task.
Upon encountering a Nestmate, the robot performs collision avoidance by adjusting its heading, and resumes its CRW search. When encountering a Non-Nestmate, the robot transitions into the \textit{Walling} state, in which it stops moving, and activates a walling timer. The timer is reset if it encounters a Nestmate in the \textit{Moving} state. Once the timer expires, the robot switches to the \textit{Avoid Non-Nestmate} state, in which it performs collision avoidance. The robot remains in this %collision avoidance 
state if its distance to Non-Nestmates is below a safe distance threshold ``SafeDist.'' It enters the \textit{Moving} state and resumes a CRW if its distance to Non-Nestmates is above SafeDist. %The timer is uniformly set to 3 seconds across both the FSM and RL-based controllers, ensuring consistent wall formation dynamics.

\subsection{Reinforcement Learning (RL) Controller} %Architecture} 
\label{sec:RL}

%\spr{(Can we combine this section and section 4 to make it clearer how the RL controller is designed?)}

%\subsubsection{Learning-based Navigation with Multi-modal Sensing}

%\spr{(Integrate somewhere?:) Learning-based Navigation with Multi-modal Sensing}

We also introduce a novel multi-agent reinforcement learning~(MARL)~\cite{OH23} approach that incorporates a DQN-based architecture as a high-level policy that selects among discrete walling, avoidance, and exploration modes, effectively learning when and where to deploy these behaviors. Furthermore, the DQN is designed for Ultra-Wideband~(UWB)-based swarm control~\cite{AlarifiEtAl16}; it processes multi-modal sensor inputs from seven nearest neighbors, providing a fixed-size representation that enables consistent inference while capturing rich spatial relationships within the swarm. For each neighbor, the network processes UWB distance measurements, Angle of Arrival~(AoA) data, and binary swarm-type indicators that distinguish between same- and opposing-swarm robots. This architecture leverages UWB's potential for accurate range measurements and combines it with deep learning techniques to create an efficient and adaptive navigation system for mobile robots.

\begin{figure}
    \centering
    \includegraphics[width=\linewidth]{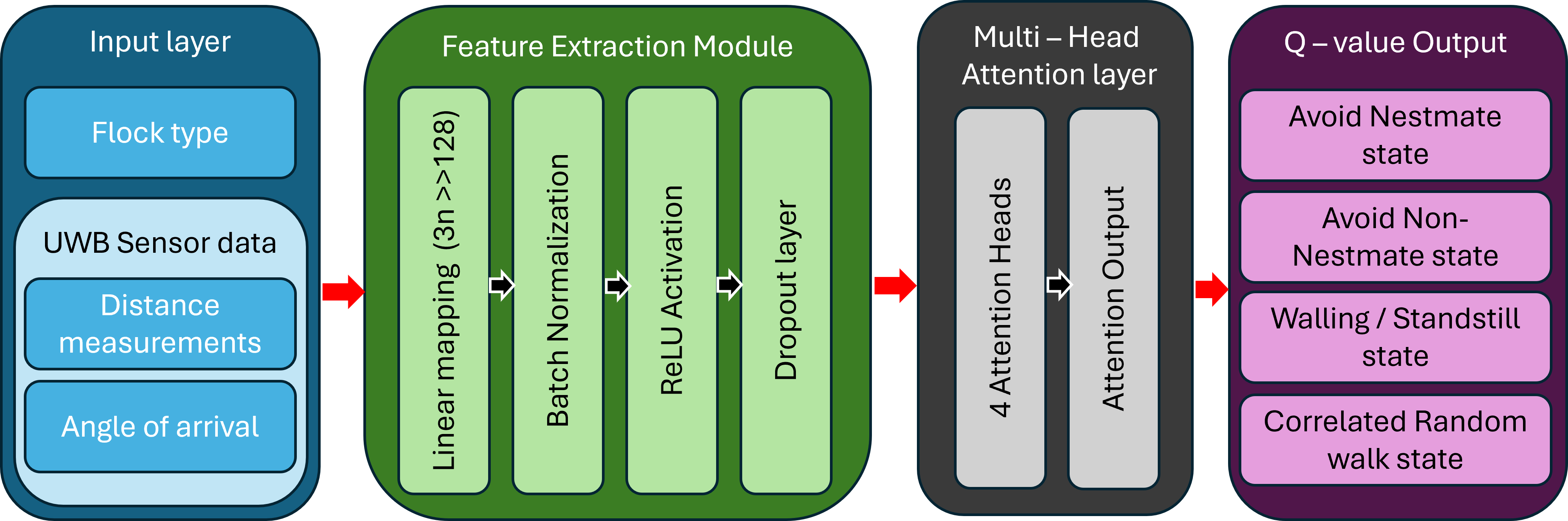}
    \caption{Reinforcement learning-based hybrid controller architecture: The %block 
    diagram illustrates the RL-based navigation system, which processes multi-modal sensor inputs from neighboring robots, including UWB distance measurements, Angle of Arrival~(AoA) data, and binary swarm-type indicators. The architecture employs a feature extraction module with a linear transformation and ReLU activation, followed by a multi-head attention mechanism for efficient processing of variable-sized input data. The final Q-value output layer generates optimal navigation actions, enabling adaptive state switching between walling, avoidance, and exploration %movement, separation, and walling 
    behaviors to maintain swarm coordination and spatial separation.}
    \label{fig:RL-block}
\end{figure}

The \textbf{proposed architecture}, shown in Fig.~\ref{fig:RL-block}, processes UWB sensor measurements to generate optimal navigation policies. 
The \textbf{input layer} accepts a variable-length sequence of UWB measurements from neighboring robots, consisting of distance measurements $d = \{d_1, d_2, ..., d_n\}$ and AoA data $\theta = \{\theta_1, \theta_2, ..., \theta_n\}$, where $n$ represents the number of observable robots within the UWB range. A binary swarm identifier is also incorporated for each neighbor.
The \textbf{feature extraction module} employs a linear transformation that maps the 3$n$-dimensional input to a 128-dimensional hidden representation, followed by batch normalization and a ReLU activation function. A dropout layer ($\text{rate} = 0.2$) enhances generalizability.
A key innovation in this architecture is the incorporation of a \textbf{multi-head attention mechanism} with four attention heads (32 dimensions each). This mechanism enables the network to process input of variable size efficiently, learn different aspects of spatial relationships simultaneously, dynamically weight the importance of different neighbors, and maintain permutation invariance in the input sequence.
The \textbf{output layer} produces Q-values for four discrete actions:
\begin{enumerate}
    \item {\it Avoid Nestmate} state (exploration)
    \item {\it Avoid Non-Nestmate} state (collision avoidance)
    \item {\it Standstill} state (position holding/walling state)
    \item {\it Random Walk} state (deadlock resolution/exploration)
\end{enumerate}

For \textbf{DQN training}, we use a centralized training with decentralized execution~(CTDE) paradigm~\cite{OH23}, which enables robust and scalable swarm navigation while maintaining computational efficiency during deployment. The environment contains two swarms of 30 robots each, with each robot perceiving its seven neighbors via UWB distances, AoA, and binary swarm-type identifiers. Training follows an episodic reinforcement learning approach, where each episode runs for 1,000 steps to ensure sufficient interactions for policy learning. The reward function balances exploration and stability while penalizing deadlocks and excessive proximity, as the RL network has access to swarm coverage and mixing ratio during training~(Section~\ref{sec:simulation}). However, collision avoidance is implemented as part of a safety mechanism, and the RL does not interact with the collision avoidance module. Notably, the network does not utilize or have access to swarm coverage or mixing ratios during inference or on individual robots. The DQN is trained using an experience replay buffer of 100,000 transitions, with a target network updated every 1,000 steps. The Adam optimizer is used with a learning rate of 0.001, batch size of 64, and discount factor of 0.99. An $\varepsilon$-greedy strategy gradually shifts from exploration to exploitation by $\varepsilon$ relaxation from 1.0 to 0.01 over 50,000 steps. Training continues for 500,000 steps, after which the learned policy is deployed in a fully decentralized manner, enabling efficient and adaptive swarm navigation. 

\section{Simulation Environment}
\label{sec:simulation}

%To validate our approach, we implemented \spr{our controllers \sout{the walling strategy}} in two distinct simulation platforms. Initial prototyping was conducted using the GIS Agent-Based Modeling Architecture (GAMA) framework~\cite{Taillandier2019}, enabling large-scale swarm simulations with 80 agents per population, each operating with a detection radius of \spr{(what does this refer to?:) 7 units.} \spr{(What are these?:)} Walling and timeout periods were set to 20 and 4 simulation steps, respectively, allowing for efficient validation of collective behaviors. 

%Under this setup, each agent performs CRW for its search strategy. Positions of agents were randomly initialized with the two populations separated. The subset of robots that encounter non-nestmates switches to the Walling state, while the other subset moves away from the walls and continues their search task. In the CRW case, there is less group cohesion and each agent will switch to the Walling state if a non-nestmate is encountered and continue walling if a nestmate is nearby to protect them from conflict. Once all encounter rates drop for some time, then the agent goes back to the Moving state. \spr{\sout{And we found that in both search strategies, utilizing walling strategy significantly delays mixing between the two populations.}}

For evaluating the SM and RL controllers, we developed a custom Pygame-based simulator~\cite{Pygame2024} % (see Fig. \ref{fig:sim}) 
for multi-robot interactions, MARL (supporting CTDE using DQN), and decentralized decision-making. The simulator models relative localization (e.g., using UWB signals), allowing each robot to track its seven nearest neighbors to support collision avoidance and local coordination.
%It includes UWB signal modeling, which allows for UWB-based localization and each robot the ability to track the distances to its seven closest neighbors (enabling collision avoidance among other things). 
Moreover, it incorporates additive noise to simulate real-world sensor uncertainties.
%To ensure safe interactions, 
%The simulator also integrates collision avoidance between robots. %\spr{(how?). \sout{, enabling robots to avoid collisions while optimizing movement. }}
For studying cooperative and adversarial behaviors among multiple co-existing swarms, the simulator realizes two distinct robot populations (swarms A \& B) that can interact in space.
%while tracking swarm coverage and mixing ratios to analyze spatial interactions.  
%Additionally, a reset simulation button is included, facilitating seamless environment resets for reinforcement learning experiments.

% Designed as a lightweight alternative to Gazebo, the simulator enables swarm robotics research, supporting walling, area coverage, UWB-based localization, and obstacle avoidance while allowing rapid prototyping and reinforcement learning policy refinement.

%% \begin{figure}[t!]
%     \centering
%     \includegraphics[width=0.7\linewidth]{Images/sim.png}
%     \caption{Custom Pygame-based simulation framework.} % for Multi-Robot Navigation}
%     \label{fig:sim}
% \end{figure}

%To evaluate swarm behavior and interaction dynamics, 
The simulator tracks \textbf{two performance metrics}: \emph{coverage} and \emph{mixing ratio}. These metrics can be used to evaluate RL %reinforcement learning 
policies and offer quantitative insight into the effectiveness of the walling strategies, collective movement patterns, and inter-swarm interactions. \vspace{2mm}

%Together, these metrics provide a robust framework for analyzing emergent swarm behaviors and evaluating reinforcement learning policies or control strategies for multi-robot coordination. 

\noindent\textbf{Coverage:}
The coverage by a swarm (A or B) is measured using the convex hull of the robots' positions, representing a measure of the total area %\spr{(clarify? it's not really the area sensed by the swarm, unless we're considering localization by triangulation)} 
occupied by the swarm. By dividing this area by the total area of the simulation environment, we obtain the percent of the environment covered by the swarm.
%This area is normalized by the total size of the simulation environment to compute the percentage of workspace covered. 
Higher coverage indicates a more dispersed spatial distribution of robots, %and area utilization, 
which is critical for tasks such as exploration, environmental monitoring, and cooperative search operations. \vspace{2mm}

\noindent\textbf{Mixing Ratio:}
The mixing ratio quantifies the degree of spatial overlap between the two swarms by computing the intersection area of their convex hulls. It is expressed as a percentage of the union of the convex hulls. A high mixing ratio indicates significant inter-swarm interaction, relevant for analyzing cooperative or competitive behaviors. Conversely, a low mixing ratio reflects spatial segregation, characteristic of effective walling formations.

%Together, these metrics provide a robust framework for analyzing emergent swarm behaviors and evaluating reinforcement learning policies or control strategies for multi-robot coordination. 

\section{Simulation Results}
\label{sec:results}

\begin{figure*}
    \centering
    \includegraphics[width=0.8\linewidth]{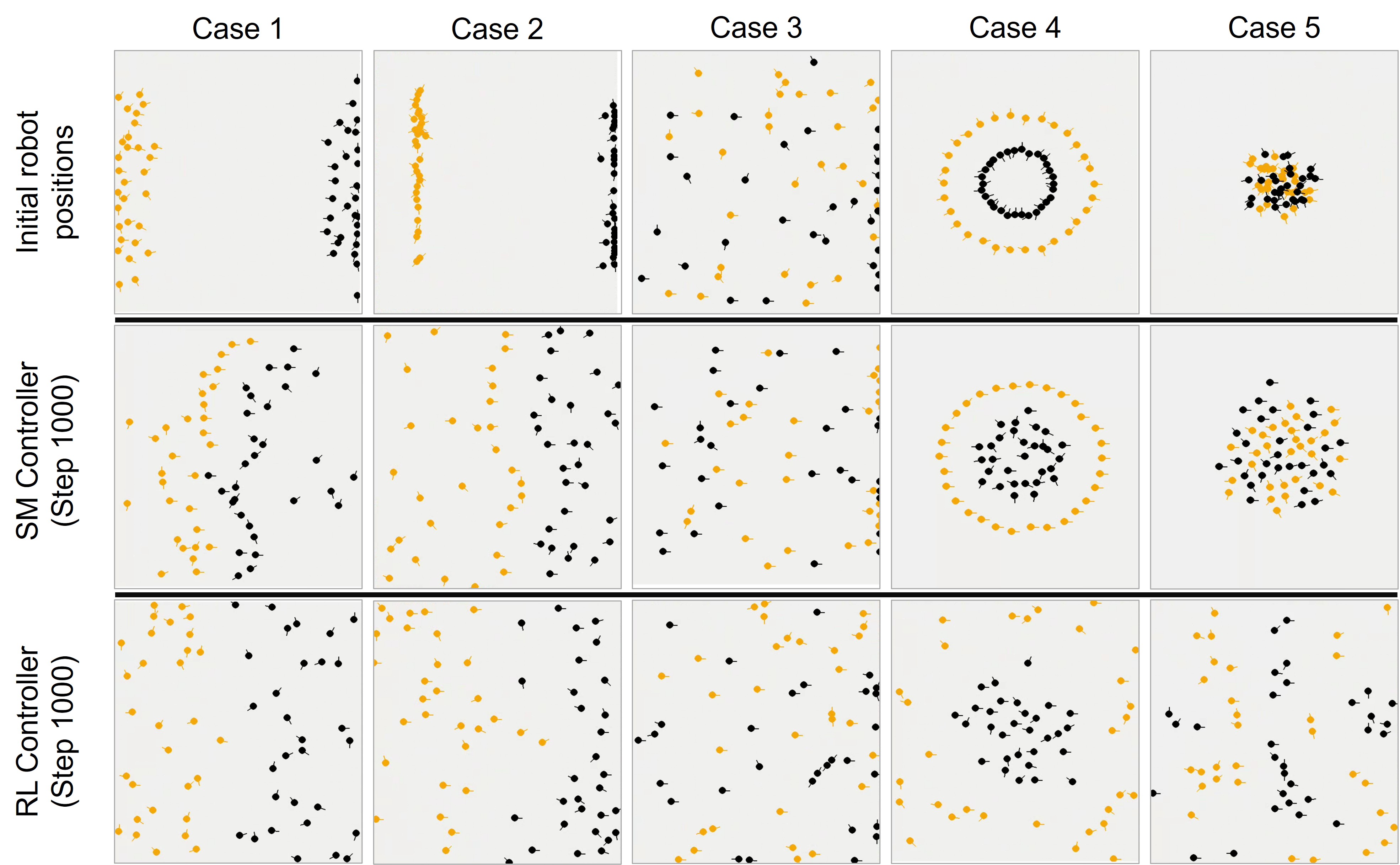} 
    \caption{\emph{Top row:}~Initial robot positions (step 1) in five simulated cases: (Case~1)~Robots initialized at the right and left edges of the environment; (Case~2)~Orange robots initialized at 1/8 the environment width from the left edge, black robots at the right edge; (Case~3)~Robots randomly initialized throughout the environment; (Case~4)~Robots initialized in concentric circles (black: inner, orange: outer); (Case 5) Robots randomly initialized near the center of the environment. \emph{Middle row:} Robot positions at step 1000 under the SM controller. %  a state machine-based controller 
    \emph{Bottom row:}~Robot positions at step 1000 under the RL controller. %a reinforcement learning (RL)-based controller 
    }
    \label{fig:eq_spaced_sim}
\end{figure*}

We evaluated the performance of our two controllers through simulations in our custom environment, comparing their coverage and mixing ratio percentages over time for five different initial swarm configurations, labeled Cases~1 through~5~(Fig.~\ref{fig:eq_spaced_sim}). To test the effect of the walling timer, we ran simulations of both controllers with this timer set either to~0 or~3 seconds. Unless indicated otherwise, results were generated from experimental treatments with 100 replications using 30 robots in each swarm; each replication ran for 5,000 time steps.  %uniformly 
%set to 3 seconds %across 
%for both %the FSM and RL-based 
%controllers, ensuring consistent wall formation dynamics.

%To validate the effectiveness of our ant-inspired walling strategies at swarm separation and to test the overall validity of reinforcement learning-based hybrid controllers, we conducted extensive simulations in our custom environment. %using \spr{(Not actual turtlebots though:)} TurtleBots. 
%We compared a simple state machine-based controller with a reinforcement learning-based hybrid controller, evaluating their performance based on coverage percentage and mixing ratio percentage. We also analyzed the impact of different initial spawning conditions and their effect on overall performance.

%%% TODO: Put all of these "Initial robot positions..." figs into a
%%%       big figure with subfigures for each of them. Would ideally be
%%5       a full-column figure. 
%%%
%%%       Also get rid of any redundant text that is shared across the 
%%%       captions and use it in a common caption for the big figure. 
%%%       Make sure each figure is labeled with a descriptive short 
%%%       title that makes it clear what it is depicting without having
%%%       to find it in the text.

%
% \begin{figure}[t]
%     \centering
%     \includegraphics[width=0.8\linewidth]{Images/g1.png}
%     \caption{Initial robot positions at step 1, with the orange robots are spawned at 1/8 from the left and the black robots at the right side of the environment (left). Robot positions at step 1000 using a state machine-based controller (center) and an RL-based controller (right).}
%     \label{fig:uneq_spaced_sim}
% \end{figure}
% %
% \begin{figure}[t]
%     \centering
%     \includegraphics[width=0.8\linewidth]{Images/g4.png}
%     \caption{Initial robot positions at step 1, with robots spawned at concentric circles, the black robot in the inner circle and the orange robot in the outer circle (left). Robot positions at step 1000 using a state machine-based controller (center) and an RL-based controller (right).}
%     \label{fig:con_sim}
% \end{figure}
% %
% \begin{figure}[t]
%     \centering
%     \includegraphics[width=0.8\linewidth]{Images/g5.png}
%     \caption{Initial robot positions at step 1, with both robots spawned at the center of the simulation (left). Robot positions at step 1000 using a state machine-based controller (center) and an RL-based controller (right).}
%     \label{fig:center_spawn_sim}
% \end{figure}
% %
% \begin{figure}[t]
%     \centering
%     \includegraphics[width=0.8\linewidth]{Images/g3.png}
%     \caption{Initial robot positions at step 1, with robots randomly spawned within the simulation (left). Robot positions at step 1000 using a state machine-based controller (center) and an RL-based controller (right).}
%     \label{fig:random_spaced_sim}
% \end{figure}
%
\begin{figure*}
    \centering
    \includegraphics[width=0.92\linewidth]{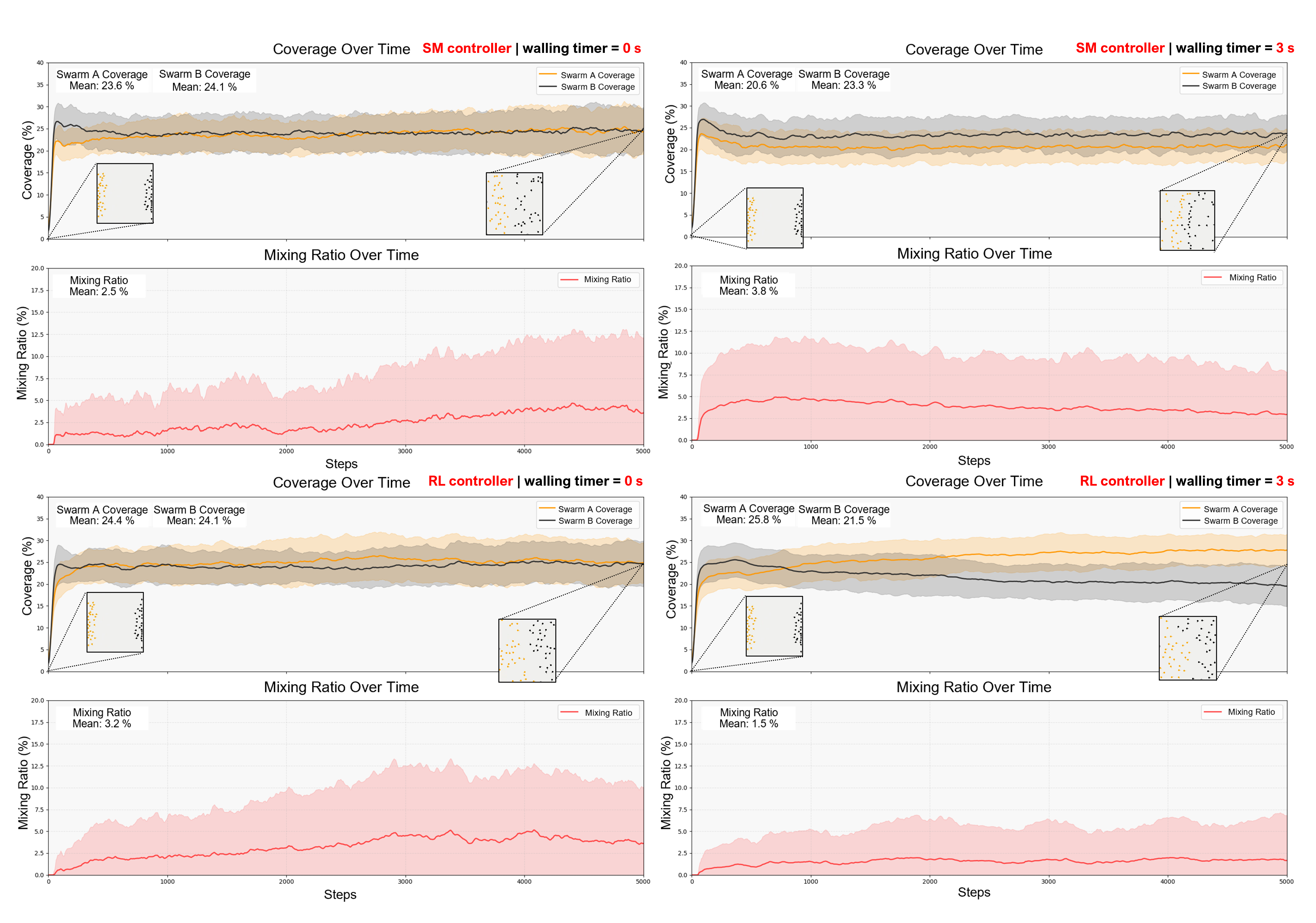}
    \caption{\textbf{Case 1:} %
    Coverage and mixing ratio percentages over time (simulation steps) for (\emph{top left}) SM controller with walling $\text{timer} = 0\,\text{s}$; (\emph{top right})~SM controller with $\text{timer} = 3\,\text{s}$; (\emph{bottom left})~RL controller with $\text{timer} = 0\,\text{s}$; (\emph{bottom right})~RL controller with $\text{timer} = 3\,\text{s}$. Solid lines (shaded regions) are stepwise averages (ranges) over 100 runs.
    %At step 1, robots are positioned in opposite sides of the environment. The plots illustrate coverage over time (coverage percentage vs. number of steps) and mixing ratio over time (mixing ratio percentage vs. number of steps) 
    %across four scenarios: (top left) State machine-based controller without a stop timer in the standstill state, (top right) State machine-based controller with a timer, (bottom left) RL-based controller without a timer, and (bottom right) RL-based controller with a timer.
    %
    % SUGGESTED CHANGE:
    %
    % Coverage percentage and mixing ratio percentage over time for both controllers. Top row: SM controller. Bottom row: RL controller. Left column: 0-second timer. Right-column: 3-second timer. Solid lines (shaded regions) are stepwise averages~(ranges) over 100 runs.
    }
    \label{fig:eq_spaced_graph}
\end{figure*}

\subsection{Qualitative Results}
The simulation snapshots in Fig.~\ref{fig:eq_spaced_sim} highlight distinct behavioral differences between the two controllers. The SM controller maintains stable wall-like structures, ensuring clear separation, but often results in rigid formations that can lead to stagnation in certain environments. In contrast, the RL controller exhibits more adaptive behavior, dynamically adjusting swarm positions to maintain separation while maximizing coverage.

When robots are initialized on opposite sides of the environment (Cases 1 and 2), both controllers successfully preserve %maintain 
the initial separation, but the RL controller converges faster, stabilizing in fewer steps. In Cases 3 and 5, the RL controller quickly disentangles the two swarms, whereas the SM controller struggles to establish clear separation due to the lack of predefined boundaries, taking significantly longer to achieve similar behaviors. A similar trend is observed in Case 4, %the concentric spawn case, 
where the RL controller breaks the concentric circle formation earlier, improving overall coverage, while the SM controller requires more time to disperse the swarms.

It is important to note that although the inclusion of a walling timer reduces coverage slightly and slows down mixing ratio convergence, it ensures to an extent that proportional coverage across the swarm is maintained. This effect is particularly evident in Case 4, %the concentric circle spawn case, 
where the structured formation remains more stable. Additionally, in Case 2, the two swarms initially achieve equal coverage, but if the orange robots are positioned closer to the center, they achieve more coverage than the black robots. This coverage imbalance is mitigated when using the controller with the nonzero walling timer, %stop timers in the standstill state, 
suggesting that the timer effectively acts as a territory marker, helping maintain proportional coverage %distribution 
across the swarms.

%%% fig:uneq_spaced_sim here

% \vspace{-5mm}

\begin{figure*}
    \centering
    \includegraphics[width=0.92\linewidth]{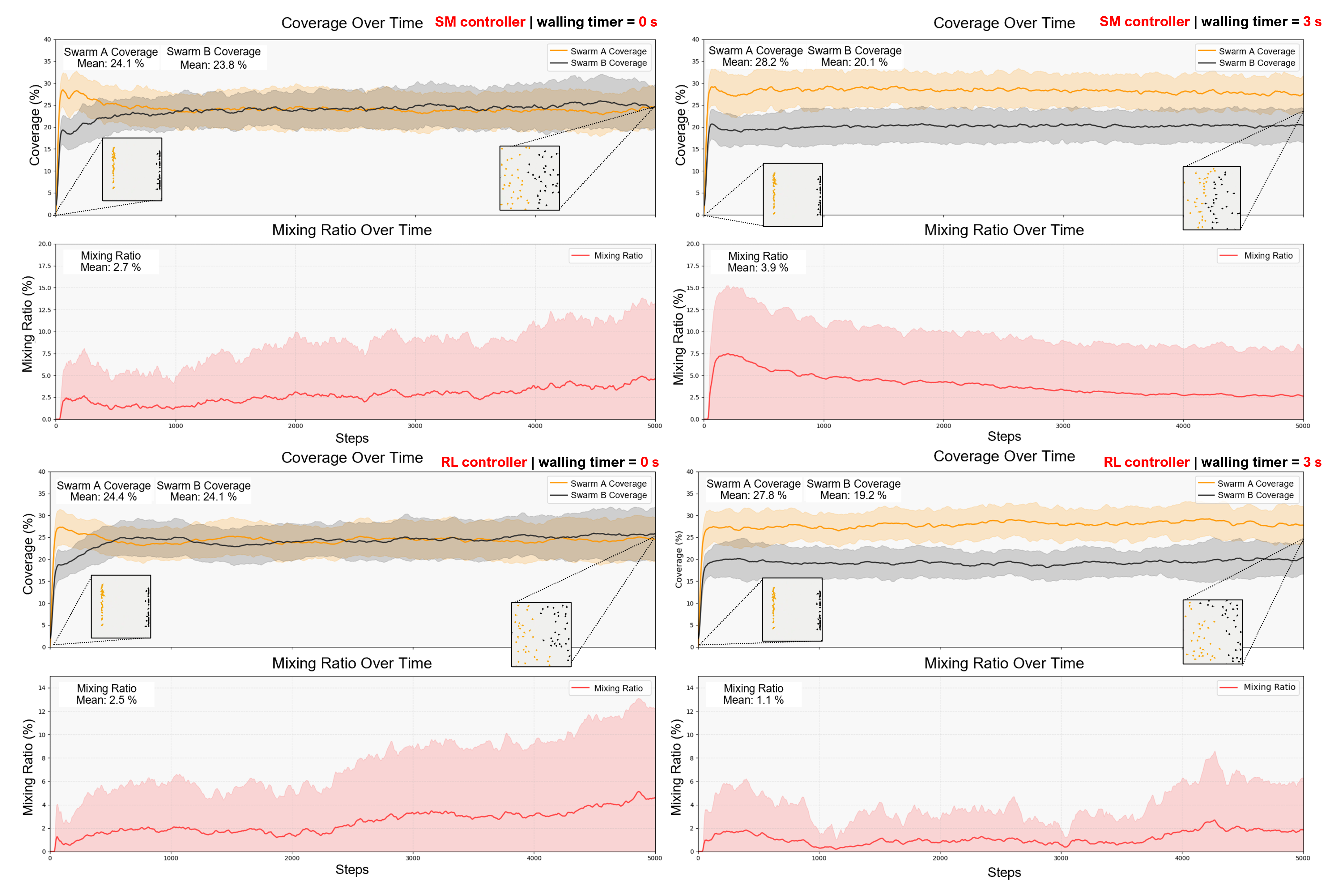}
    \caption{\textbf{Case 2:} %
    Coverage percentage over time and mixing ratio percentage over time for (\emph{top left})~SM controller with walling $\text{timer} = 0\,\text{s}$; (\emph{top right})~SM controller with $\text{timer} = 3\,\text{s}$; (\emph{bottom left})~RL controller with $\text{timer} = 0\,\text{s}$; (\emph{bottom right})~RL controller with $\text{timer} = 3\,\text{s}$. Solid lines (shaded regions) are stepwise averages (ranges) over 100 runs.}
    \label{fig:uneq_spaced_graph}
\end{figure*}

%%% fig:random_spaced_sim here

\begin{figure*}
    \centering
    \includegraphics[width=0.92\linewidth]{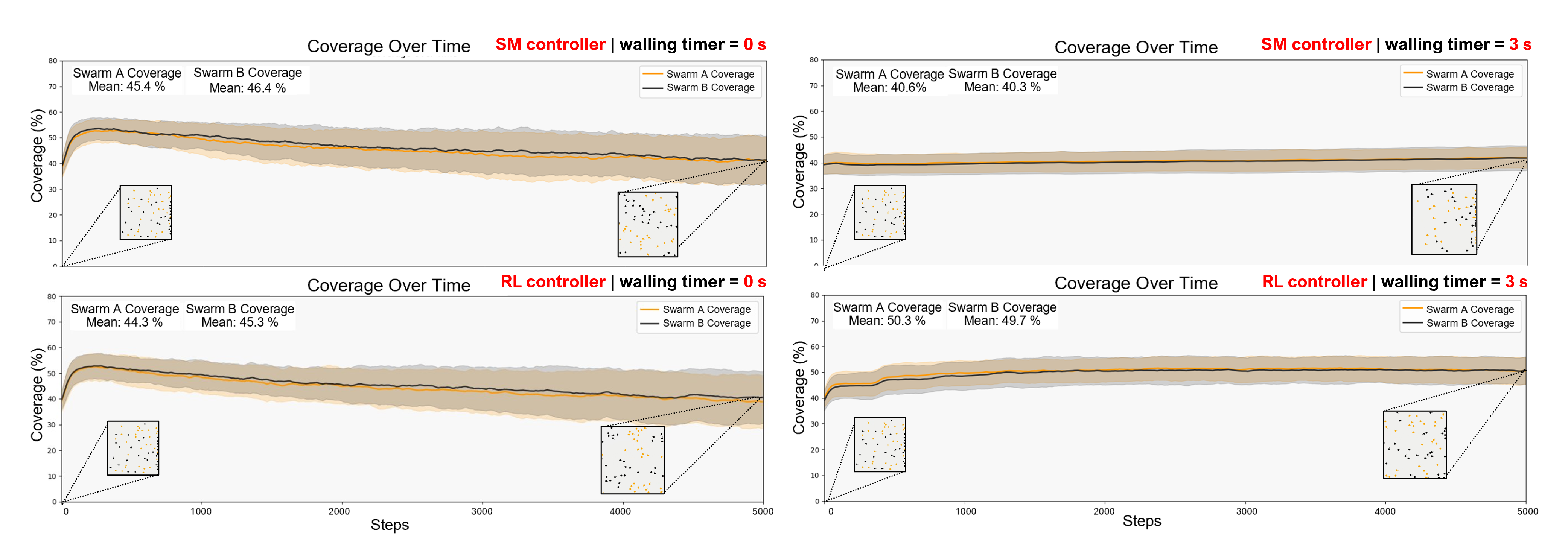}
    \caption{\textbf{Case 3:} %
    Coverage percentage over time for (\emph{top left})~SM controller with walling $\text{timer} = 0\,\text{s}$; (\emph{top right})~SM controller with $\text{timer} = 3\,\text{s}$; ({\it bottom left}) RL controller with $\text{timer} = 0\,\text{s}$; (\emph{bottom right})~RL controller with $\text{timer} = 3\,\text{s}$. Solid lines (shaded regions) are stepwise averages (ranges) over 100 runs.}
    \label{fig:random_spaced_graph}
\end{figure*}

\subsection{Quantitative Results}

\begin{figure*}
    \centering
    \includegraphics[width=0.92\linewidth]{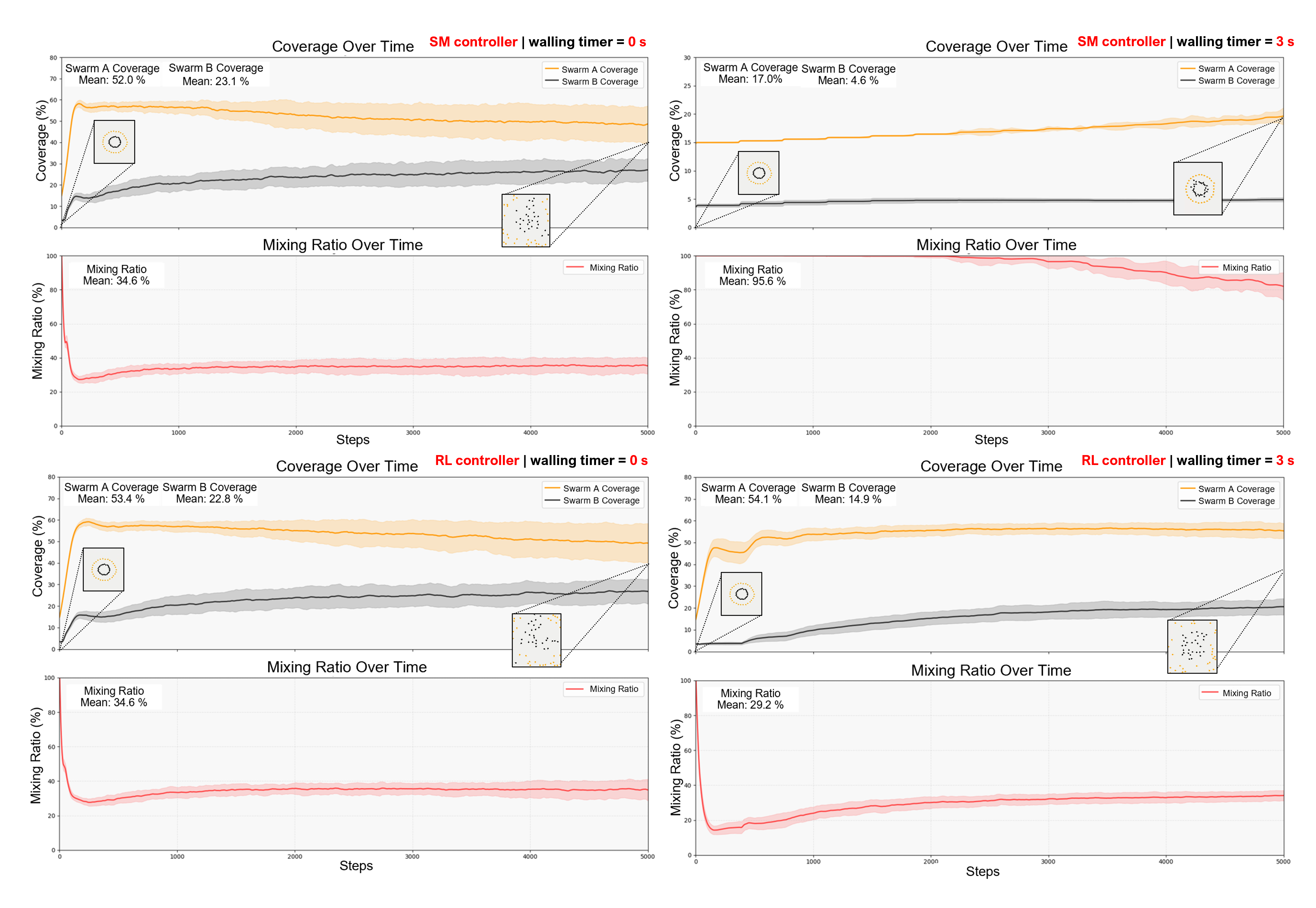}
    \caption{\textbf{Case 4:} %
    Coverage percentage over time and mixing ratio percentage over time for (\emph{top left})~SM controller with walling $\text{timer} = 0\,\text{s}$; (\emph{top right})~SM controller with $\text{timer} = 3\,\text{s}$; (\emph{bottom left})~RL controller with $\text{timer} = 0\,\text{s}$; (\emph{bottom right})~RL controller with $\text{timer} = 3\,\text{s}$. Solid lines (shaded regions) are averages (ranges) over 100 runs.}
    \label{fig:con_graph}
\end{figure*}

\begin{figure*}
    \centering
    \includegraphics[width=0.92\linewidth]{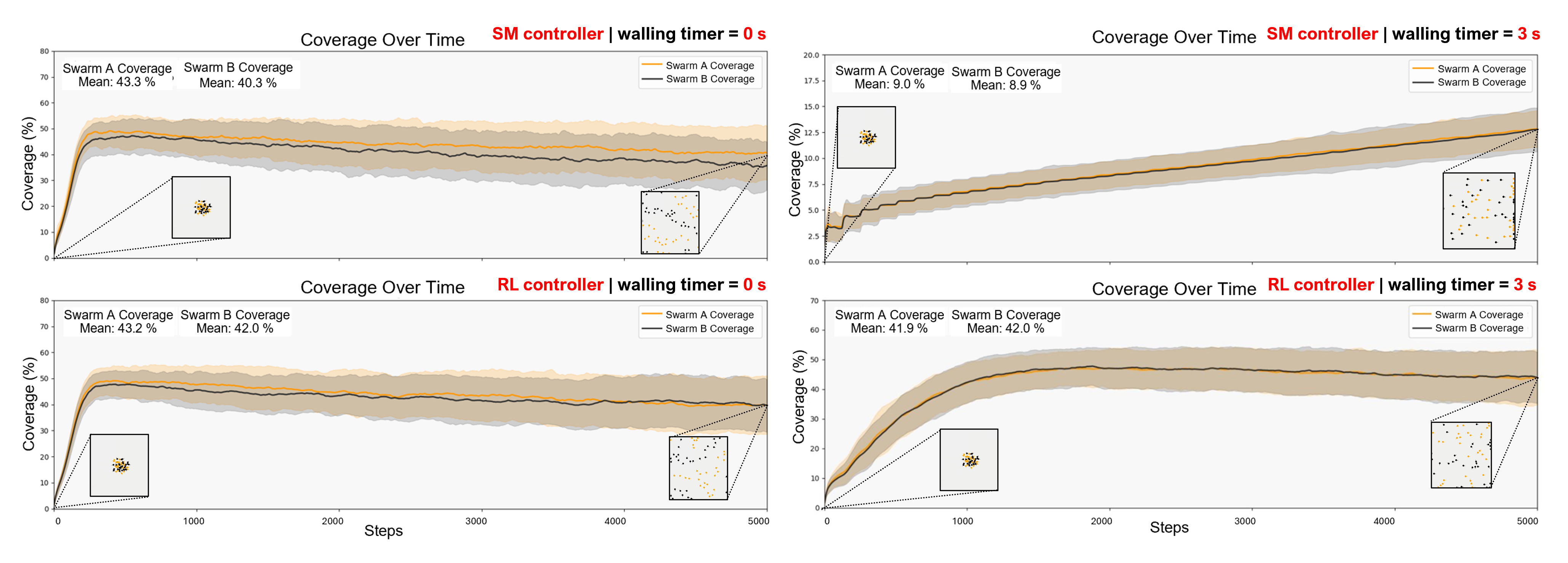}
    \caption{\textbf{Case 5:} %
    Coverage percentage over time for (\emph{top left})~SM controller with walling $\text{timer} = 0\,\text{s}$; (\emph{top right})~SM controller with $\text{timer} = 3\,\text{s}$; (\emph{bottom left})~RL controller with $\text{timer} = 0\,\text{s}$; (\emph{bottom right})~RL controller with $\text{timer} = 3\,\text{s}$. Solid lines (shaded regions) are averages (ranges) over 100 runs.}
    \label{fig:center_spawn_graph}
\end{figure*}

Figures~\ref{fig:eq_spaced_graph}--\ref{fig:center_spawn_graph} show the coverage percentage and, where nonzero, mixing ratio percentage for Cases 1--5 with both controllers and the walling timer set to either~0 or~3~s. They demonstrate 
%The quantitative evaluation of the proposed hybrid RL-based state machine (SM) controller demonstrates its 
the effectiveness of the RL controller at maintaining spatial separation between different types of robots and optimizing coverage while preserving the structured decision-making of traditional state machines. Across the cases, %multiple experimental conditions, including opposite-side, random, concentric, and centralized spawning scenarios, 
the RL controller %RL-based FSM 
exhibited superior adaptability in reducing mixing ratios and optimizing spatial coverage compared to purely state-machine-driven approaches. Specifically, in Case 1, %the opposite-side spawn case, 
the RL controller stabilized in fewer than 250 simulation steps, while the SM controller required 400--500 steps to reach a comparable state. In Case 3, the RL controller achieved a mixing ratio reduction of 40--50\% relative to the SM controller within the first 300 steps, indicating a more efficient disentanglement of the two swarms. The incorporation of walling timers %in the standstill state 
was observed to enhance proportional coverage distribution, mitigating biases introduced by initial swarm positioning.

\begin{figure}
    \centering
    \includegraphics[width=\linewidth]{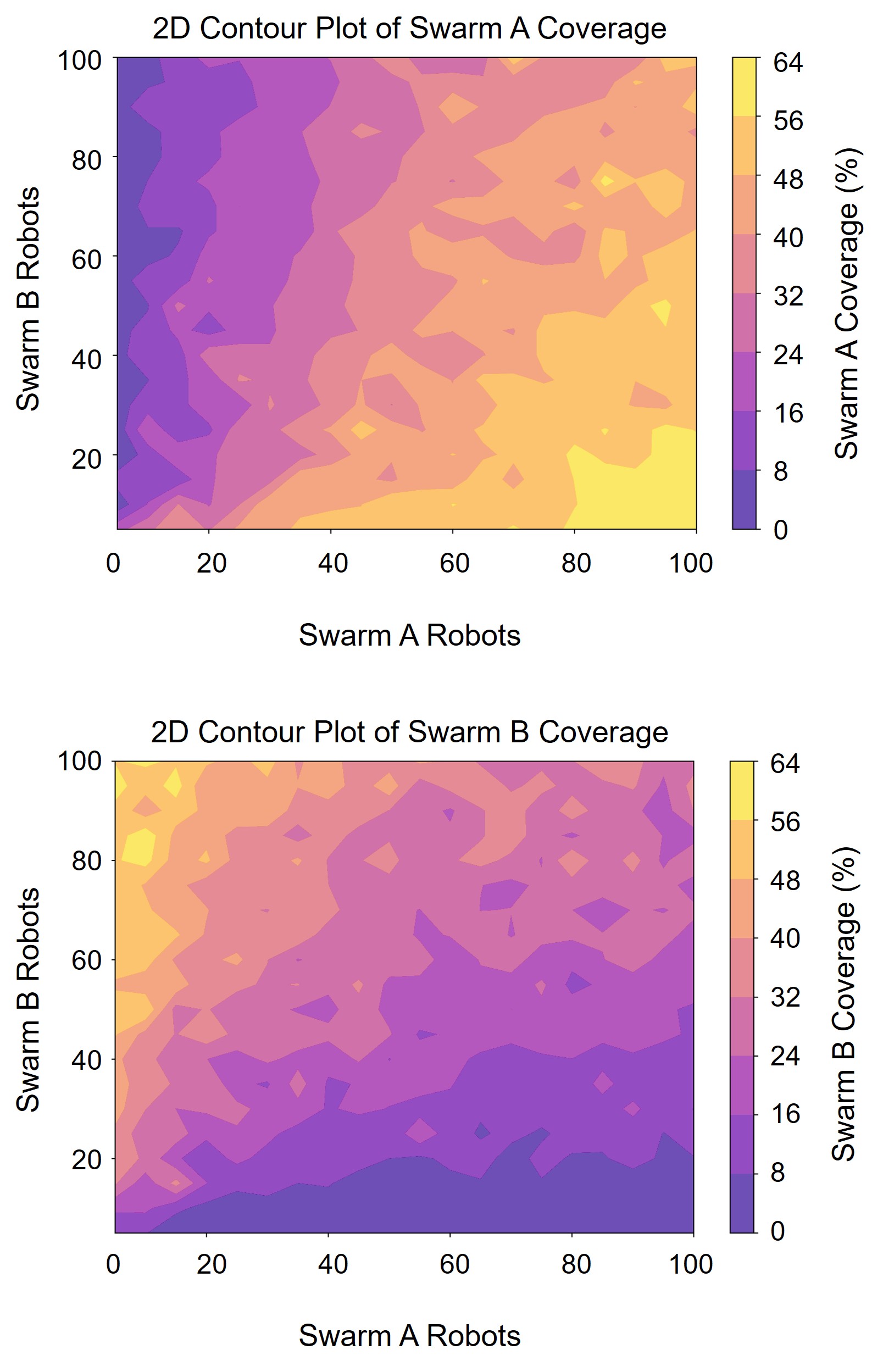}
    \caption{\textbf{Case 1:} \emph{(Top:)}~Coverage percentage by swarm A vs.~population sizes of swarm A and swarm B, and \emph{(bottom:)}~coverage percentage by swarm B vs.~population sizes of swarm A and swarm B. Coverages are averaged over two runs.}
    \label{fig:multiple}
\end{figure}

\begin{figure}
    \centering
    \includegraphics[width=\linewidth]{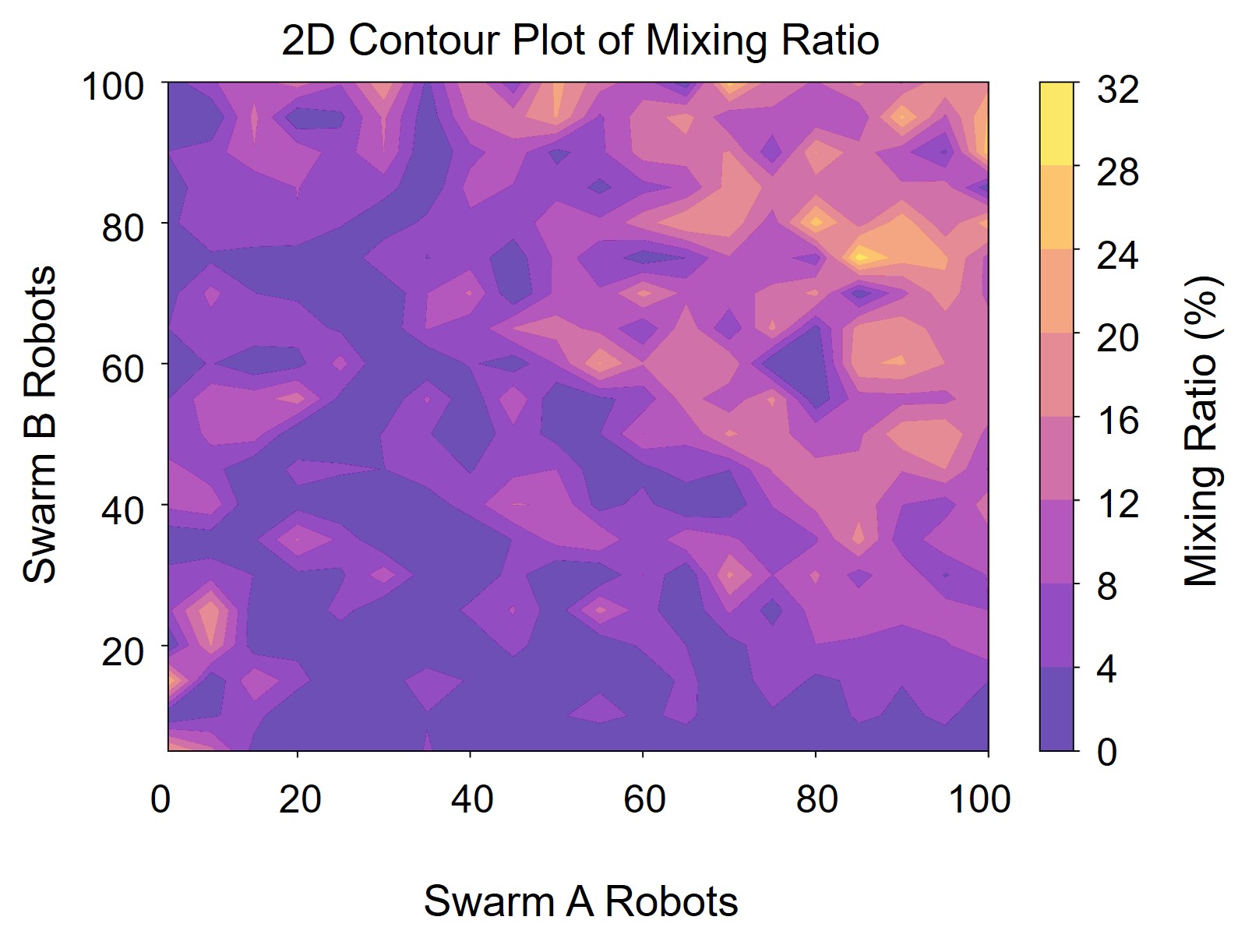}
    \caption{\textbf{Case 1:} Mixing ratio percentage vs. population sizes of swarm A and swarm B. Ratios are averaged over two runs.}
    \label{fig:multiple_mix}
\end{figure}

Furthermore, as shown in Fig.~\ref{fig:multiple}, evaluations across swarm sizes ranging from 10 to 100 robots per swarm indicate that the RL controller maintains consistent separation and coverage performance, provided that a minimum number of robots are available to sustain effective walling structures. Notably, Fig.~\ref{fig:multiple_mix} shows a significant spike in the mixing ratio when the number of robots per swarm falls below 15, regardless of the opposing swarm’s size, highlighting a critical threshold for maintaining continuous wall formations. This suggests that below this minimum population density, the swarm lacks sufficient robots to establish stable barriers, leading to increased inter-swarm mixing and reduced separation efficacy.

These results show that the RL controller
%emphasize that the RL-based state machine 
is not merely an RL-based alternative to traditional state machines, but rather an effective optimization framework that enhances state transitions and adaptive decision-making within a state-machine structure, offering a scalable and data-driven mechanism to improve swarm autonomy while retaining the interpretability of finite-state logic.

%%% fig:con_sim here

% \vspace{-5mm}

%%% fig:center_spawn_sim here

% \vspace{-5mm}

\section{Conclusion \& Future Work}
\label{sec:conclusion}

This work presents an ant-inspired walling strategy for swarm robotic systems, demonstrating how a reinforcement learning (RL)-based state machine with Deep Q-Networks~(DQN) can serve as a scalable and adaptive alternative to traditional finite state machines~(FSMs). Although state-machine-based controllers have been widely used in swarm robotics for their reliability and simplicity, they often require extensive manual tuning and can struggle in unstructured or dynamically changing environments. Our results suggest that augmenting state machines with RL, particularly using DQN-based decision-making, provides a flexible and effective approach to decentralized swarm coordination.  

By integrating DQN-based learning with a structured FSM framework, our approach maintains the interpretable and modular nature of state machines while incorporating adaptive behaviors that improve coverage, reduce swarm mixing, and enhance scalability. The ability of RL-enhanced state machines to autonomously adjust behaviors based on local sensing and environmental conditions makes them a compelling alternative to fully end-to-end deep reinforcement learning methods such as PPO~(Proximal Policy Optimization) or SAC~(Soft Actor--Critic), which often require extensive computational resources and complex reward engineering. 

Our findings highlight that swarm robotic systems do not need to fully abandon state-machine-based architectures, but can instead leverage RL-driven state switching to enhance decision-making and adaptability without introducing unnecessary complexity. This approach is particularly valuable for real-world deployments where computational efficiency, explainability, and robustness to sensor noise are critical.  

%Future work can further refine 
This method can be further refined by incorporating multi-agent communication, heterogeneous swarm roles, and %real-world 
testing on physical robots. Ultimately, RL-based state machines with DQN provide a practical, scalable, and adaptable solution for decentralized swarm robotics, enabling self-organizing behaviors without the computational overhead of complex deep RL architectures.

\section{Acknowledgment}

This work was funded in part by NAVAIR contract N6833522G093. The views and conclusions contained in this document are those of the authors and should not be interpreted as representing the official policies, either expressed or implied, of the Naval Air Warfare Center or the U.S. Government.

%%%%%%%%%%%%%%%%% BIBLIOGRAPHY IN THE LaTeX file !!!!! %%%%%%%%%%%%%%%%%%%%%%
% \begin{thebibliography}{9}
% \bibitem{ref1}
% DARS-SWARM2020 Website,\\ ``https://www.swarm-systems.com/dars-swarm2020''

% \bibitem{ref2}
% T. Mure, ``XXX Control of Nonlinear Systems'',
% {\it International Journal of YYY}, Vol. 99, No. 99, pp. 123--456, 2012.

% \bibitem{ref3}
% H. Gun, {\it ZZZ Handbook}, SICE, Tokyo, 2012.

% \bibitem{ref4}
% T. Mure, ``Measurement Method using ABC Sensor'', {\it Transaction of EFG}, 
% Vol. 00, No. 1, pp. 1--9, 2011.

% \bibitem{ref5}
% H. Gun, ``SSS Intelligent Systems'', {\it International Journal of VVV}, 
% Vol. 12, No. 5, pp. 500--600, 2011.

% \end{thebibliography}

\bibliographystyle{IEEEtran}
\bibliography{walling}

\end{document}